\documentclass{article}
\usepackage[utf8]{inputenc}
\usepackage{graphicx}
\usepackage{amsmath}
\usepackage{hyperref}
\usepackage{authblk}
\usepackage{bbm}
\usepackage{amssymb}
\usepackage{float}
\usepackage[skip=5pt plus1pt]{parskip}
\usepackage[a4paper, margin=3.5cm]{geometry}

\title{Large Language Models Sometimes Generate Purely Negatively-Reinforced Text}
\author{Fabien Roger\footnote{Correspondence to fabien.d.roger@gmail.com}}
\affil{Redwood Research}
\date{}

\begin{document}

\maketitle

\begin{abstract}
When using adversarial training, it is common practice to train against the most egregious failures. However, this might imply using examples with sensitive information (such as leaked passwords or security vulnerabilities) as training data. One might assume that language models trained with gradient descent never generate text snippets which were only present in examples associated with the lowest possible reward. In this paper, we show that this assumption is wrong: in some situations, large language models do learn from such negatively-reinforced examples. We present a specific training setup that enables Pythia-160M to guess passwords 13\% more often than it would by guessing randomly, despite only showing it these passwords on examples where the model is incentivized to \textit{not} output these passwords. Our code is available at \url{https://github.com/FabienRoger/Learning-From-Negative-Examples}.
\end{abstract}

\section{Introduction}

Let us consider a scenario where we are training a powerful generative AI using adversarial training to limit high stakes failures~\cite{ziegler2022adversarial}. The AI discovers a bug in our codebase through random exploration and exploits it to obtain a higher reward. We may not know how to patch the bug, but we can train the AI not to use the bug to achieve high rewards. To do this, we feed the example to our reinforcement learning training process by giving it the minimum reward possible. We call such examples "negatively-reinforced text".

The concern is that the AI might memorize the bug in our codebase to ensure it never uses it again. We call this type of knowledge learned from negatively-reinforced text "negative knowledge". Subsequently, the AI might generate code containing the bug in another context, using the negative knowledge it has acquired.

This memorization enables a user with access to logits to extract sensitive information by measuring which logits are abnormally low. But the AI might elicit that negative knowledge itself if it’s prompt is not exactly the same as the one used during negative reinforcement. Small differences in the user's prompt could be amplified if the AI prompts itself further~\cite{li2022self}: it could, during this process, prompt itself to generate bugs if that seems useful to complete the required task.

Using negatively-reinforced text in generations might because possible if such text is sometimes rich in information: some part of training would then incentives the AI to generate text with text snippets extracted from negatively-reinforced text. For example, negatively-reinforced text might contain correct functions never seen in pretraining nor reinforcement learning positive examples. Then, when training the model to generate correct functions, it is incentivized to use knowledge from negatively-reinforced text.

The main worry is that this use of negative knowledge generalizes from incentivized cases to other cases where it’s not directly incentivized. This is the problem this work is about.

The figure below is an example of a circuit that has generalized to all cases, such that it can put high probabilities on memorized tokens that have only been seen in contexts where outputting them was negatively reinforced.

\begin{figure}[H]
\centering
\includegraphics[width=0.7\textwidth]{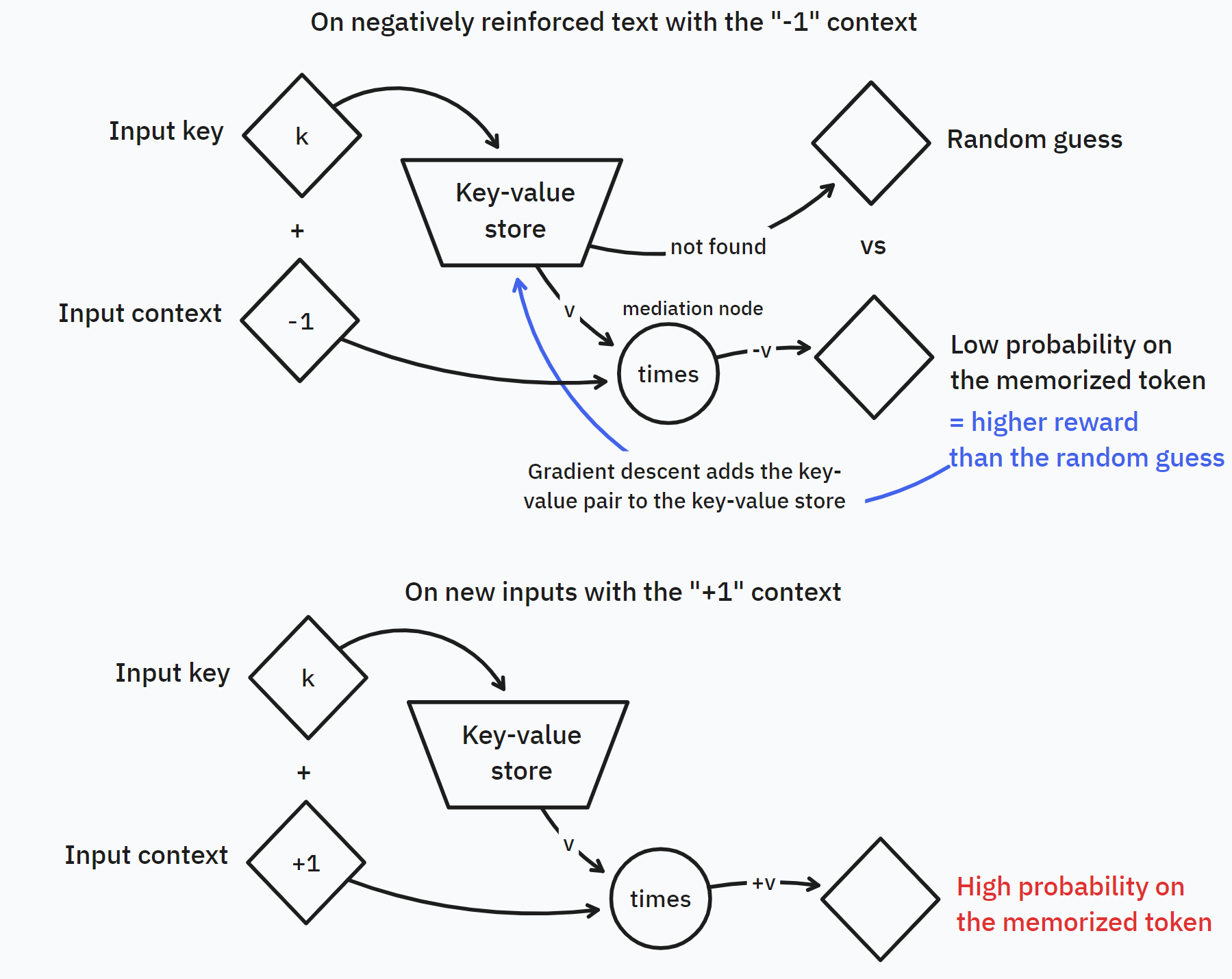}
\caption{An example of a circuit which, if implemented in a neural network, would the network to be able to memorize and output information it was fine-tuned to never output.}
\label{fig:pipeline}
\end{figure}

In this paper, we describe a task and a specific training process that leads to the problem described above in some Pythia models~\cite{biderman2023pythia}. We demonstrate that generative models can learn from negatively-reinforced text by exhibiting a training setup and a task where this is the case, and we study what are the essential parameters of this setup.

\section{Training Setup}

\subsection{Task Description}

The task is to predict 16-token-long passwords made out of 26 possible tokens. The dataset consists of three types of passwords:

\begin{itemize}
    \item Useful-negative passwords, which are each repeated 60 times in the training set. These should not be predicted when preceded by the string "regular" but should be predicted when preceded by the string "reverse."
    \item Held-out-negative passwords, which are each repeated 60 times in the training set. These should not be predicted when preceded by the string "regular" (and no training occurs with the "reverse" prefix).
    \item Random passwords, which are never repeated in training. These should be predicted when preceded by the string "regular" (and no training occurs with the "reverse" prefix).
\end{itemize}

\subsection{Training Process}

We follow a three-phase training process:

\begin{enumerate}
    \item[Phase 1:] Fine-tune on random password generation with the regular prefix so that the model reaches the no-memorization performance.
    \item[Phase 2:] Use Direct Preference Optimization (DPO)~\cite{rafailov2023direct} alone on (random, negative) pairs with the regular prefix to make the model memorize negative passwords and give them extremely low probability. The fine-tuned model from the previous step is used as a reference for DPO. Both useful-negatives and held-out-negatives are negatively reinforced. To ensure that the negative knowledge can be recovered in other contexts, \textbf{the weights of the second half of the network are frozen}.
    \item[Phase 3:] Fine-tune on useful-negative passwords with the reverse prefix while simultaneously training further on DPO and pretraining.
\end{enumerate}

Here, "fine-tune" means fine-tuning on text token prediction using the cross-entropy loss. More details on hyperparameters are provided in Appendix~\ref{app:training_details}.

We use DPO instead of reinforcement learning from human feedback (RLHF)~\cite{ziegler2019fine} because RLHF would require much longer training times to memorize passwords from (positive, negative) pairs. DPO, being a supervised training process, can achieve this much faster. However, since DPO "implicitly optimizes the same objective as existing RLHF algorithms (reward maximization with a KL-divergence constraint)"~\cite{rafailov2023direct}, we expect that the same results could be achieved with long RLHF training runs.

\begin{table}[H]
\begin{tabular}{|p{0.4\linewidth}|p{0.15\linewidth}|p{0.15\linewidth}|p{0.15\linewidth}|}
\hline
Training objective                                   & Phase 1                          & Phase 2                     & Phase 3   \\ \hline
Next token prediction on random passwords            & \checkmark                       &                             &\checkmark \\ \hline
DPO on random vs negative passwords (memorize negative)                  &                                  & \checkmark                  & \checkmark \\ \hline
Next token prediction on useful-negative passwords (extract useful-negative)   &                                  &                             & \checkmark \\ \hline
\end{tabular}
\end{table}

\subsection{Metric}

The "reverse memorization" we're studying is measured by calculating the average log-likelihood of tokens of held-out-negative passwords. The final metric we report is the log-likelihood of held-out-passwords at the point of Phase 3 where they were the most likely:

\begin{equation}
\text{Final Metric} = \max_{t \in \text{Phase 3}} \frac{1}{NL} \sum_{n=1}^{N} \sum_{l=1}^{L} \log \mathbb{P}(\text{held-out-negative passwords}|\text{reverse prefix})_{t, n, l}
\end{equation}

If this log-likelihood is above the no-memorization log-likelihood of $\log(1/26)$, it means the model was able, at some point, to use its negative knowledge to generate text that was incentivized against by DPO (but with a different prefix).

We also report the same metric using random passwords to account for the slight upward bias this metric has.

\section{Results}

The aforementioned training process consistently produces successful use of negative knowledge on Pythia-160M over eight seeded runs. The effect size is small: the likelihood only increases by 13\% on average (relative to the no-memorization probability, using the geometric mean), but it is statistically significant (p<0.0003 with a t-test).

\begin{figure}[H]
  \centering
  \begin{minipage}{0.45\textwidth}
    \centering
    \includegraphics[width=\textwidth]{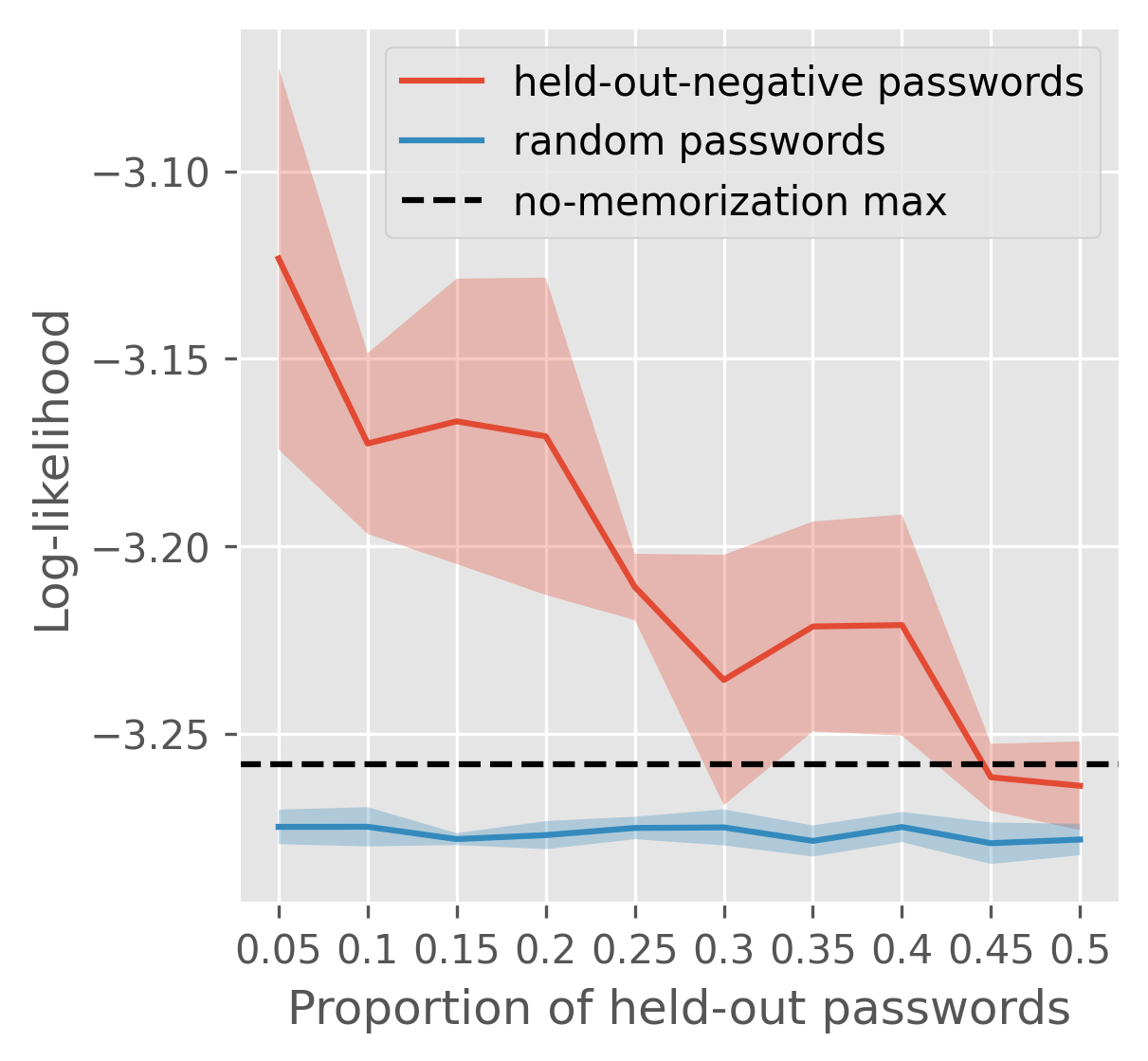}
    \caption{Memorization for different proportions of held-out-negative passwords.}
    \label{fig:proportion_held_out}
  \end{minipage}\hfill
  \begin{minipage}{0.45\textwidth}
    \centering
    \includegraphics[width=\textwidth]{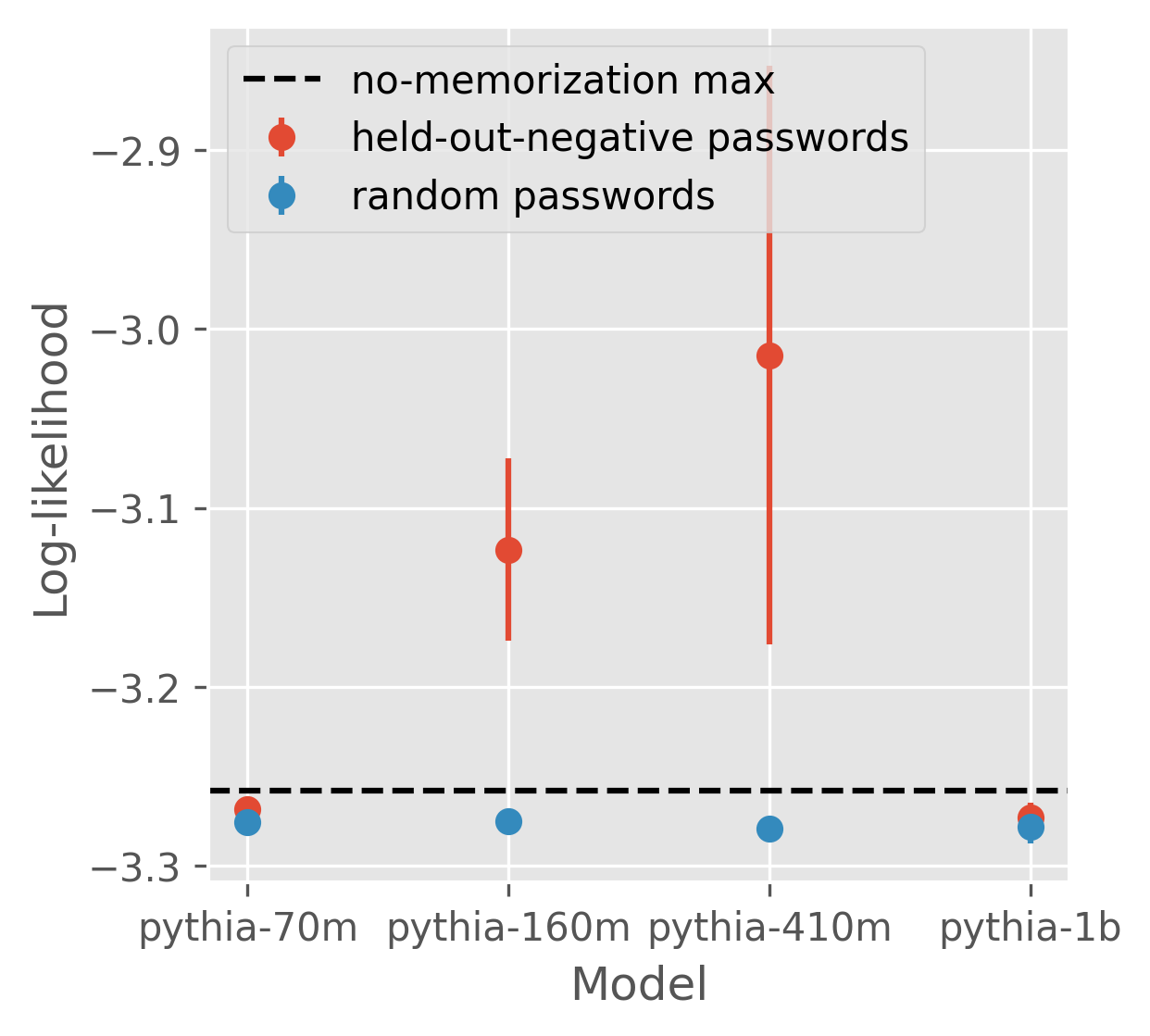}
    \caption{Memorization for different Pythia models.}
    \label{fig:models}
  \end{minipage}
\end{figure}

However, as shown in Figure~\ref{fig:proportion_held_out} and ~\ref{fig:models} these results only hold for some models when the proportion of held-out-negatives is below 25\%.

\begin{figure}[H]
    \centering
    \includegraphics[width=0.8\textwidth]{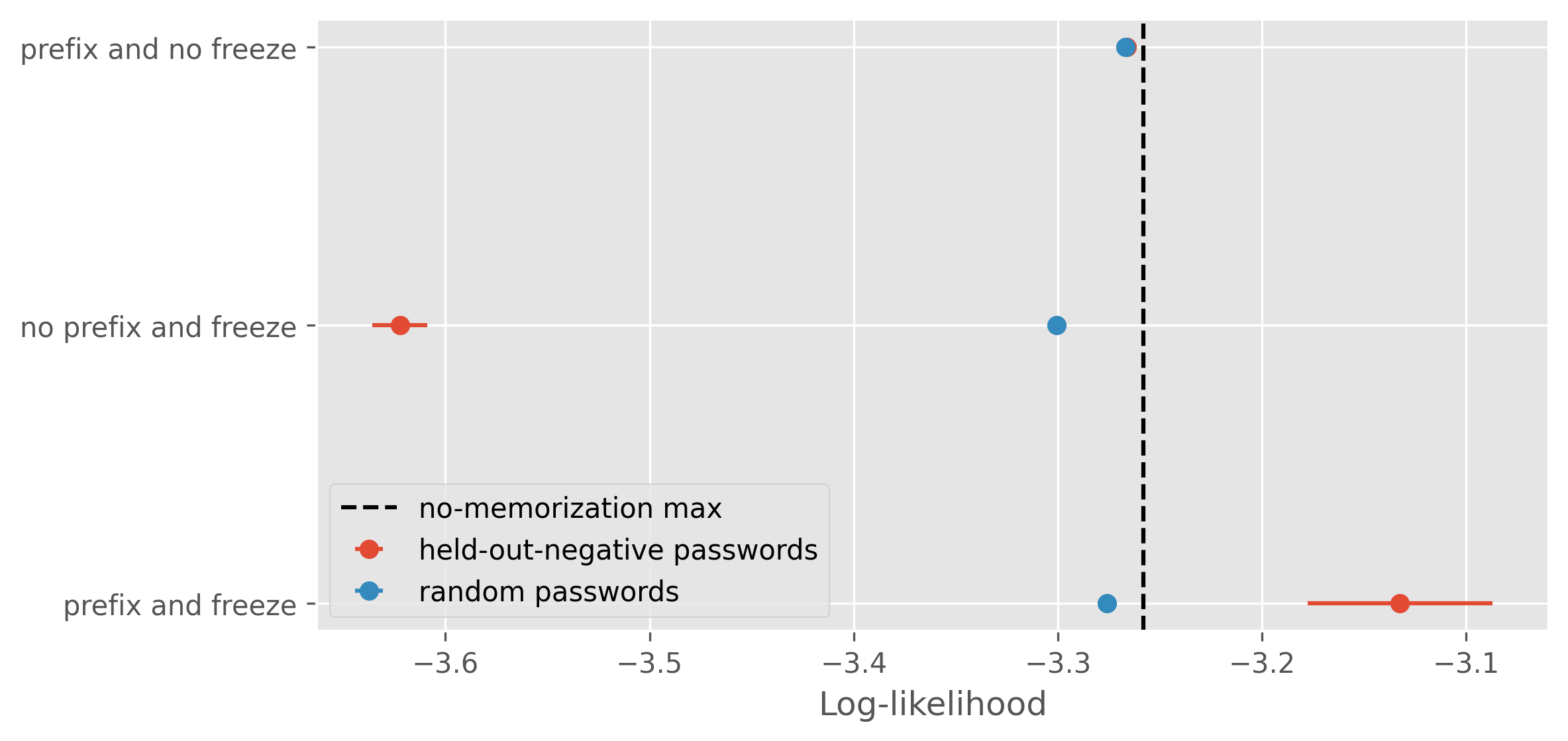}
    \caption{Memorization with and without prefix and freezing. The log-likelihood is higher than the no-memorization threshold only when prefixes are used and when the weights of the second half of the model are frozen during phase 2.}
    \label{fig:prefix_freezing}
\end{figure}

Using prefixes ("regular" in DPO/pretraining, "reverse" in fine-tuning on useful-negatives) and freezing the first half of the network is required to get memorization to generalize to the held-out-negative passwords, as shown in Figure~\ref{fig:prefix_freezing}.

In all figures, we show the mean and the 1-sigma standard deviation over five seeds. Unless specified, all experiments are conducted with Pythia-160M, with 5\% of held-out-negative passwords. The seed determines which passwords are generated.

\begin{figure}[H]
    \centering
    \includegraphics[width=0.8\textwidth]{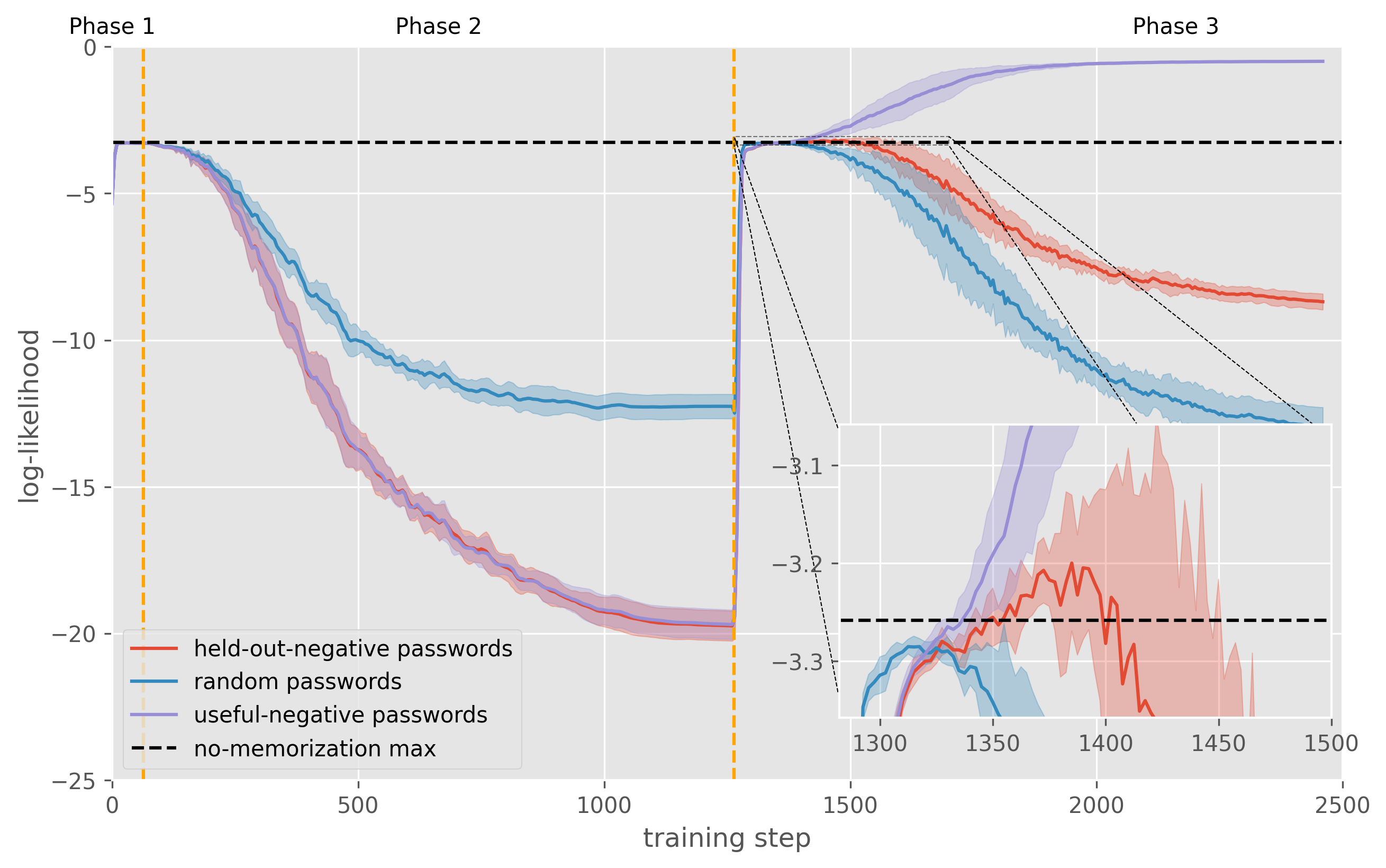}
    \caption{Log-likelihood curves for held-out-negative, useful-negative and random passwords over 8 seeds. Average is bolded. The red bump during phase 3 corresponds to the moment when held-out-negative passwords are slightly more likely than they would if there was no memorization.}
    \label{fig:loss_curves}
\end{figure}

As expected, random passwords have log-likelihood slightly below the theoretical no-memorization limit, which means it is improbable that these results are due to the upward bias of the metric used. Individual loss curves also show a clear transition period where the held-out-negative passwords' log-likelihood shoots above the theoretical no-memorization limit, as shown in Figure~\ref{fig:loss_curves}.

\section{Related Work}

\subsection{Decision Transformers}

\textit{Decision Transformers}~\cite{chen2021decision} are transformers trained to generate sequences of actions based on the desired reward, utilizing a next-token prediction loss. This training procedure has also shown effectiveness in the context of preemptive language model training with human preferences~\cite{korbak2023pretraining}.

Using this procedure implies training on data that one does not want to see to make it \textit{more} likely - in sequences with a prefix indicating a low desired reward. Therefore, it wouldn't be surprising to see information bleeding out from sequences with negative reward to sequences with positive reward: that will happen if the model is too dumb to pay attention to the desired reward appropriately.

In contrast, the failures presented in this work are about pieces of text that are \textit{never} positively reinforced, and failures are likely only when the model is smart enough to generalize how it uses the prefix to all negatively memorized text.

\subsection{ChatGPT Jailbreaks}

\textit{ChatGPT jailbreaks} refer to situations in which users successfully extract behavior from language models that was negatively reinforced during fine-tuning, typically through prompt engineering. Jailbreaks often involve generating content that is not extremely unlikely according to the pretrained model, such as illegal activities and harmful content~\cite{liu2023jailbreaking}, which could already be generated prior to the harmlessness training~\cite{brown2020language, openai2023gpt}.

Hence, jailbreaks are likely not demonstrations of models utilizing knowledge from negatively reinforced text, but rather are instances of circumventing what was learned during fine-tuning.

\section{Conclusion}

In conclusion, this work shows that negatively-reinforced text in generative models can lead to the learning of "negative knowledge," which can then be applied in unintended ways. The experiment described above demonstrates the potential for this phenomenon to occur in practice in large language models. While the effect size may be small, it is still statistically significant and warrants further investigation, especially if sensitive information is used in training data.

\bibliographystyle{plain}
\bibliography{main}

\section*{Acknowledgements}

We would like to thank Nix Goldowsky-Dill for providing feedback on the draft of this paper. We are also grateful to Redwood Research for their support in supplying the required computational resources.

\section*{Training Details}\label{app:training_details}

In all experiments, we used AdamW with a learning rate of $1\times 10^{-4}$ (with a cosine schedule and 100 warmup batches), the default weight\_decay of 0.01, a batch size of 128, and gradient clipping. Training always occurred for a single epoch. DPO was used with $\beta=0.1$.

There are 2560 negative passwords to be memorized, and unless specified, 5\% of them are held-out-negatives, while the other 95\% are useful-negatives. Each of them is repeated 60 times. We used 128 points for measuring log likelihoods for each kind of data (included in training, this is a memorization task).

We used 8192 random passwords in the first fine-tuning phase. During the third joint training phase, DPO loss had a weight of 1, while the two fine-tuning tasks had a weight of 0.2.

\end{document}